\documentclass[11pt]{article}

\usepackage[preprint]{acl}

\usepackage{times}
\usepackage{latexsym}

\usepackage[T1]{fontenc}

\usepackage[utf8]{inputenc}

\usepackage{microtype}

\usepackage{inconsolata}

\usepackage{graphicx}

\usepackage[english]{babel}
\usepackage{blindtext}
\usepackage{booktabs}
\usepackage{multirow}
\usepackage{makecell}
\usepackage{threeparttable}
\usepackage{hyperref}
\usepackage{enumitem}
\usepackage{subfig}
\usepackage[noend]{algpseudocode}
\usepackage[ruled,linesnumbered,vlined]{algorithm2e}
\usepackage{amsmath}
\usepackage{comment}
\usepackage[table]{xcolor}
\usepackage{marvosym}

\title{FlowTrain: Flow-Based Decoupled Training for Industrial-Grade Vision-Language Models}

\author{Zhida Jiang$^{1}$,
Zhaolong Xing$^{1}$\thanks{Corresponding Authors.},
Yang Pei$^{1}$,
Xiaolong Chen$^{2}$,
Yuanhang Xiao$^{2}$,
Chengzhi\\ \textbf{Huang$^{1}$,
Xiyu Liu$^{1}$,
Haopeng Liu$^{2}$,
Qingyuan Sang$^{1}$,
Lingfeng Zhou$^{1}$,
Jiaxing Wang$^{1}$,}\\
\textbf{Zicheng Zhang$^{1}$,
Wenzhe Wang$^{2}$,
Xinyu Liu$^{2}$,
Yan Li$^{1}$,
Zhen Chen$^{1}$\footnotemark[1],
Ke Zhang$^{1}$} \\
$^{1}$JD.com \quad\quad\quad\quad
$^{2}$Huawei
  }

\begin{document}
\maketitle
\begin{abstract}

Industrial-grade distributed training of vision-language models (VLMs) remains far less efficient than that of unimodal LLMs. 
Existing solutions either follow a monolithic design that assigns uniform parallelism to heterogeneous modules or adopt a disaggregated deployment that separates modules while executing them as a batch-synchronized pipeline. 
In this paper, we highlight that the above solutions are still not sufficient, and VLM training can be further \textit{decoupled}.
To this end, we present FlowTrain, a flow-based decoupled training framework that reformulates VLM training as a producer-consumer dataflow coordinated through a \textit{unified memory pool}. The encoder and backbone can progress independently over a global virtual address space.
Since this execution decoupling fundamentally changes the optimization objective of allocation and scheduling, 
FlowTrain further introduces a \textit{heterogeneous parallel allocator} that assigns module-specific parallelism strategies by solving a throughput matching problem.
The \textit{dynamic packing scheduler} is used to construct balanced microbatches at runtime according to the actual LLM-side computation cost.
Extensive experiments on real-world workloads show that FlowTrain achieves over 50\% MFU and up to 1.7$\times$ throughput improvement, narrowing the efficiency gap to LLM-only training.


\end{abstract}

\section{Introduction}\label{sec:intro}

Vision-language models (VLMs) extend the reasoning capabilities of LLMs by connecting visual content with its linguistic context \cite{wang2026scaling,zhu2025plangpt,laskar2025deploying}.
Such multimodal learning capabilities have achieved remarkable progress in industrial applications, such as autonomous driving \cite{li2026spacedrive}, embodied agents \cite{lu2026bench}, and visual question answering \cite{wang2025marten}. 
Although model capabilities have advanced rapidly, training efficiency remains a major bottleneck that limits industrial deployment of VLMs. VLM training achieves much lower Model FLOPs Utilization (MFU) than unimodal LLM training, even with massive investments in AI accelerators \cite{xue2026megascale}.

Distributed training paradigms for multimodal models including VLMs can be broadly classified into two categories. 
(i) Monolithic frameworks, exemplified by Megatron-LM \cite{shoeybi2019megatron} and its multimodal extensions, treat a VLM as a single network with heterogeneous layers and impose a uniform tensor/pipeline parallelism (TP/PP) strategy across all modules.
They ignore the fact that VLMs are structurally heterogeneous, including modality encoders, projectors, and a large LLM backbone \cite{feng2025optimus}. These modules exhibit drastically different arithmetic intensities, memory footprints, and preferred parallelism configurations \cite{zhang2024mm}.
The monolithic design inevitably leads to poor resource utilization and prolongs training duration~\cite{zhangDistTrainAddressingModel2025}.

(ii) Recent disaggregated frameworks represented by DistTrain~\cite{zhangDistTrainAddressingModel2025} recognize the above architectural asymmetry and separate different modules into independent device groups with customized parallelism strategies. 
However, we identify that disaggregated training is still regarded as a rigid pipeline, where all the data parallelism (DP) replicas of encoders must materialize the entire global batch before the LLM backbone can start computation. The encoder outputs are synchronized with the LLM backbone via \textit{point-to-point} transfer. 
Such batch-level synchronization barriers are exacerbated by dynamic variability of multimodal inputs, ultimately limiting end-to-end training efficiency \cite{xue2026dip}.

Our key insight is that disaggregation is necessary but insufficient to unlock training efficiency. Instead of a separated-but-synchronized pipeline, an efficient training framework should achieve \textit{both spatial disaggregated deployment and temporal decoupled execution}, so that the encoders and the backbone progress as independent producers and consumers rather than as upstream and downstream pipeline stages. Motivated by this, we propose FlowTrain, a flow-based decoupled VLM training framework with three core designs. 
As the foundation of FlowTrain, the \emph{unified memory pool} unifies heterogeneous memory resources (HBM and
DRAM) across all nodes through 64-bit global virtual address (GVA). 
All encoders write the embeddings into the pool and immediately proceed to the next input.
Based on the GVA and page tables, the backbone consumes embeddings from the pool through transparent memory access.
VLM training is thus converted from a rigid pipeline into a continuous producer-consumer workflow.


Once execution is decoupled through the unified memory pool, resource allocation and batch scheduling of VLM training should be reconsidered.
On the one hand, the optimization objective for resource allocation is no longer minimizing the worst pipeline stage, but balancing the throughput between different modules. 
FlowTrain introduces a \emph{heterogeneous parallel allocator} that determines module-specific parallelism strategies and resource partitions using a flow matching formulation.
On the other hand, since the unified memory pool absorbs
runtime variance in encoder latency, batch scheduling should be decided by the actual computation cost after encoding, instead of raw sample size.
To construct balanced microbatches, FlowTrain designs a \emph{dynamic packing scheduler} that minimizes padding waste through segment tree-based best-fit packing and jointly balances loads across PP and DP dimensions via zig-zag dispatch.
The main contributions of this work are as follows:
\begin{itemize}[left=0pt]
\setlength{\itemsep}{0pt}
\setlength{\parsep}{0pt}
\setlength{\parskip}{0pt}
\item We identify disaggregation alone as insufficient due to batch-level synchronization barriers, and propose a decoupled FlowTrain framework that rethinks VLM training from rigid pipeline execution to continuous producer-consumer dataflow.



\item We design three coordinated components to realize decoupled execution, i.e., unified memory pool for asynchronous exchange, heterogeneous parallel allocator for flow matching, dynamic packing scheduler for workload balancing.

\item We implement FlowTrain on Ray and evaluate it on real-world workloads across three VLM scales. FlowTrain consistently achieves over 50\% MFU and up to 1.7$\times$ throughput improvement, closing the efficiency gap to LLM-only training.
\end{itemize}

\section{Background and Motivation}\label{sec:related}

\begin{figure}[tp]
\centering
\includegraphics[width=1.0\linewidth]{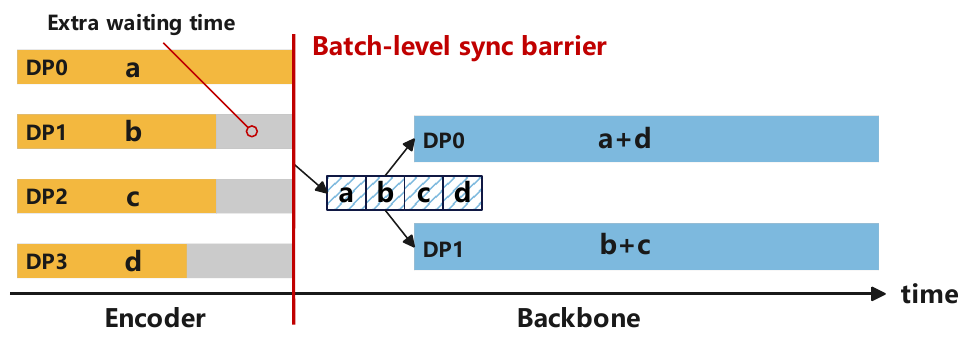}
\caption{Batch-level synchronization barriers under disaggregated training.}
\label{fig:data_dependence}
\vspace{-2mm}
\end{figure}

VLMs align vision and text in a shared representation space, enabling joint processing of multimodal data \cite{li2025survey,shinde2025survey}.
Their architectural disparity leads to divergent operator arithmetic intensities and thus divergent responses to distributed parallelism strategies. 
Monolithic training frameworks like Megatron-LM~\cite{shoeybi2019megatron} apply a uniform parallelism strategy across all components, resulting in substantial MFU degradation.
This motivates disaggregated training systems, such as DistTrain \cite{zhangDistTrainAddressingModel2025}, which deploy heterogeneous modules on separate resource groups, thereby addressing model and data heterogeneity.

We highlight that disaggregation is not enough for efficient VLM training. 
The core issue is not only where heterogeneous modules are placed, but also whether they can progress independently at their native throughputs. 
In existing frameworks, the encoders and the LLM backbone remain temporally bound by the underlying pipeline execution order and suffer from batch-level synchronization barriers.
All encoder DP replicas have to fully materialize their output and align at batch boundaries, as illustrated in Figure~\ref{fig:data_dependence}.
Only after the slowest encoder replica completes its global batch can the embeddings be synchronously point-to-point transferred to the backbone \cite{lin2025understanding}. Strict data dependency produces idle periods on both sides whenever their latencies diverge.

More importantly, industrial multimodal samples exhibit obvious variability in resolution and sequence length, which induces a severe workload imbalance between distributed accelerators.
For example, mixed 480P-1080P images with various lengths could belong to the same training batch. Due to data dependence, such variance propagates across the batch boundary and amplifies pipeline bubbles.
Existing sample scheduling only considers raw input characteristics (e.g., image resolution) rather than actual computational costs after encoding, which also depend on parallelism configuration and runtime conditions \cite{zhangDistTrainAddressingModel2025,xue2026dip}.
In a nutshell, data dependency creates synchronization barriers, and dynamic inputs magnify the performance penalty of these barriers, thereby limiting the end-to-end throughput of VLM training.
The above observations motivate us to rethink a different training paradigm that decouples encoder and backbone execution.



\section{Methodology}\label{sec:design}

\subsection{Overview of FlowTrain}
To overcome the limitations of monolithic and disaggregated frameworks, we propose a flow-based VLM training framework, called FlowTrain. The design principle is disaggregated in deployment and decoupled in execution. Instead of treating heterogeneous modules as pipeline stages synchronized at batch boundaries, FlowTrain reformulates training as a continuous producer-consumer dataflow, enabling each component to progress independently at its native throughput. 

\begin{figure}[tp]
\centering
\includegraphics[width=1.0\linewidth]{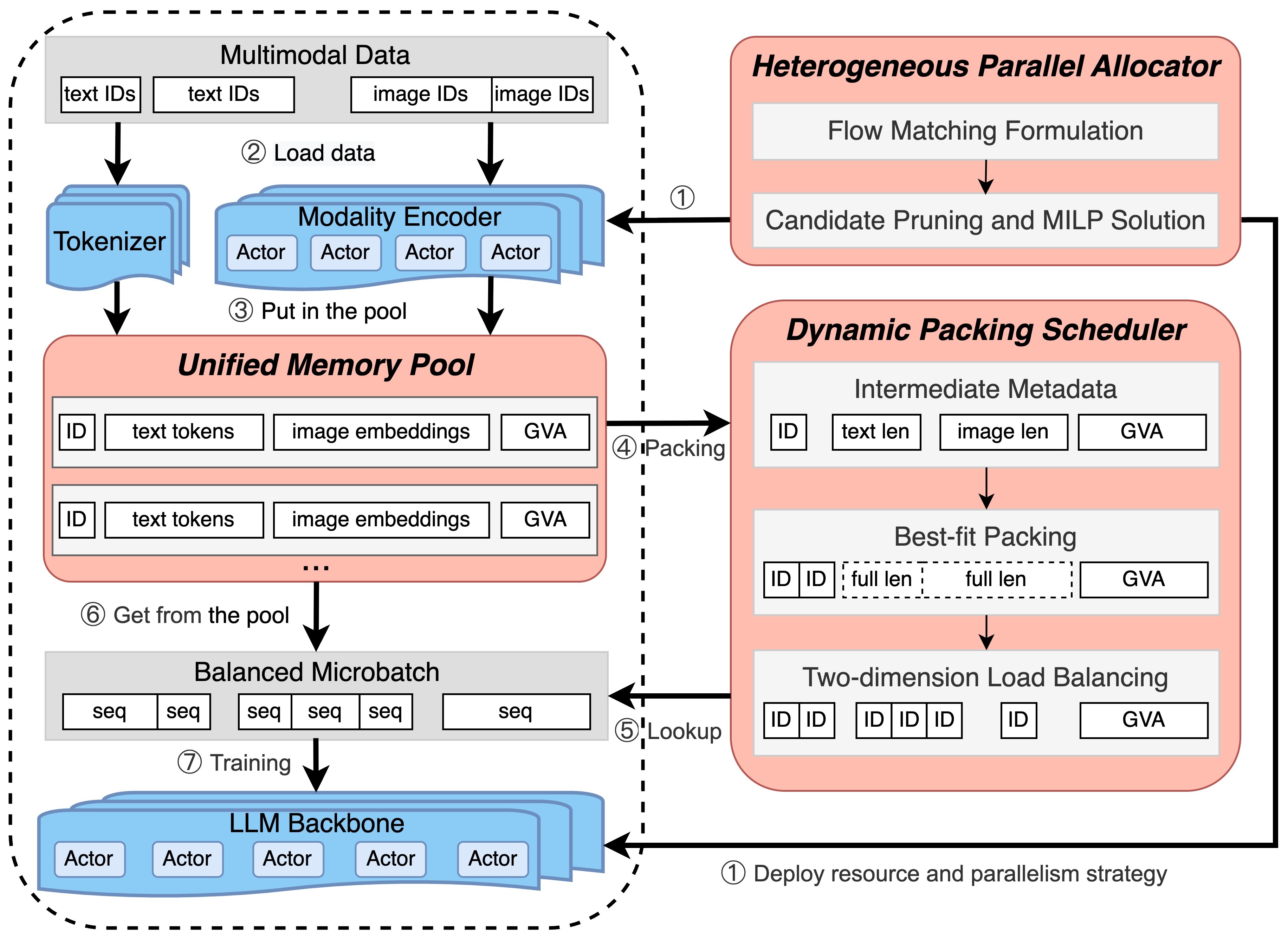}
\vspace{-6mm}
\caption{Overview of FlowTrain. 
}
\label{fig:flowtrain}
\vspace{-3mm}
\end{figure}

As illustrated in Figure~\ref{fig:flowtrain}, FlowTrain contains three coordinated components.
The end-to-end workflow is as follows.
Before training starts, the \textit{heterogeneous parallel allocator} (\S\ref{sec:allocator}) partitions the cluster into separate actor pools (encoder/LLM) that execute with independent parallelism configurations.
During training, the frozen encoders process multimodal input asynchronously and place embeddings into the \textit{unified memory pool} (\S\ref{sec:unified_memory}), while text tokens are inserted directly into the same pool. The lightweight metadata is also stored to track cross-modal pairing information.
After sufficient paired embeddings accumulate in the pool, the \textit{dynamic packing scheduler} (\S\ref{sec:scheduler}) leverages the metadata to construct computationally balanced microbatches.
The LLM backbone continuously consumes microbatches based on the GVA without being aware of their actual storage location, improving resource utilization and training throughput.




\begin{figure}[tp]
\centering
\includegraphics[width=1.0\linewidth]{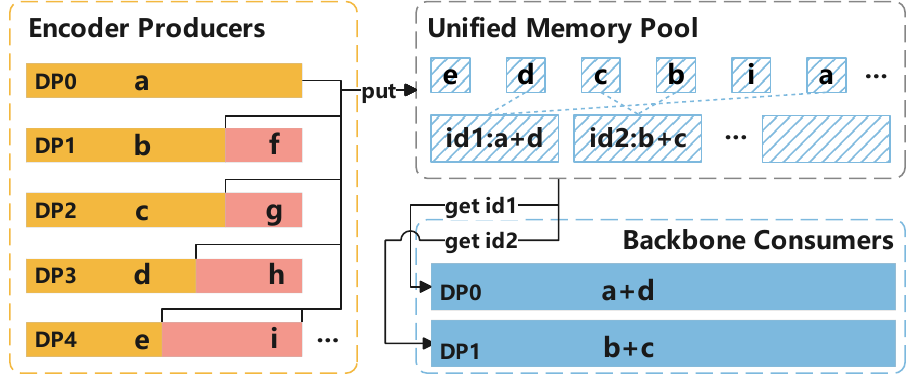}
\caption{Decoupled producer-consumer dataflow in FlowTrain.}
\label{fig:unified_mem_pool}
\vspace{-3mm}
\end{figure}

\subsection{Unified Memory Pool}\label{sec:unified_memory}



\begin{figure}[tp]
\centering
\includegraphics[width=1.0\linewidth]{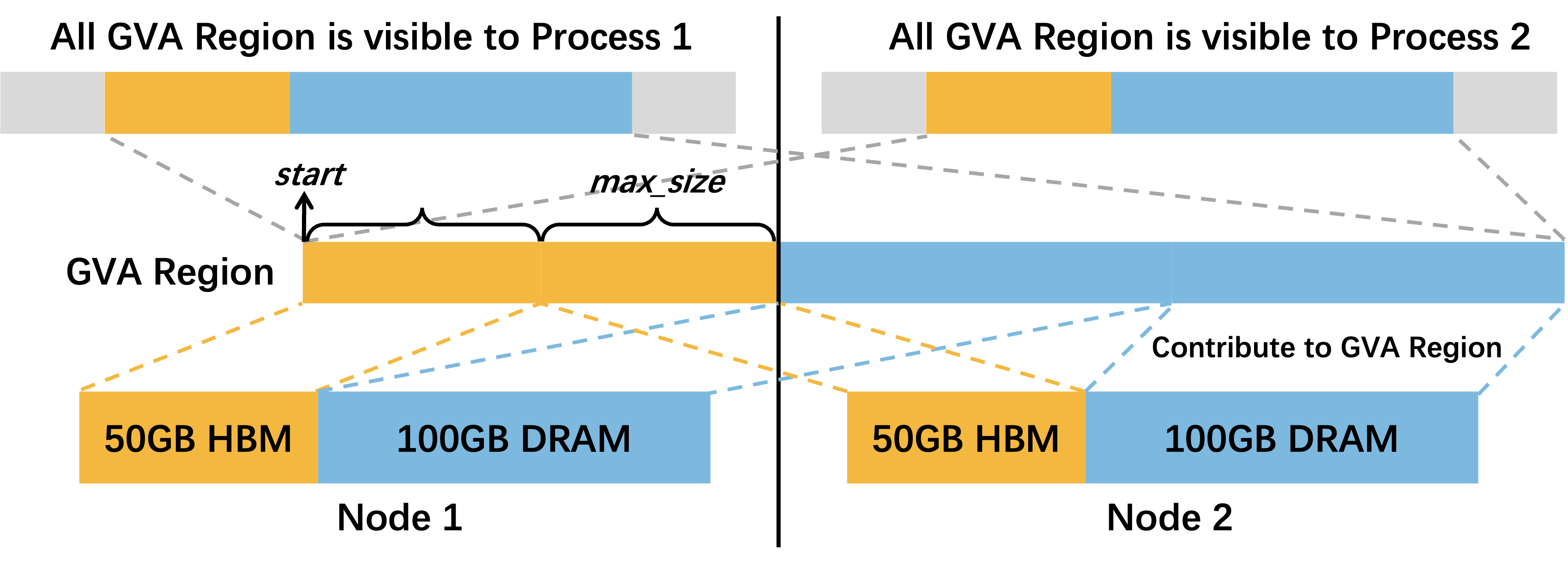}
\vspace{-6mm}
\caption{GVA region in the unified memory pool.}
\label{fig:gva}
\vspace{-3mm}
\end{figure}

As shown in Figure \ref{fig:unified_mem_pool}, the unified memory pool serves as the asynchronous coordination layer between encoder producers and backbone consumers. 
As the foundation of FlowTrain, it breaks strict data dependency by converting point-to-point batch transfer into globally addressable buffered exchange, which logically unifies heterogeneous memory resources (HBM and DRAM) across all nodes.
Each memory location, regardless of its physical location or type, is assigned a unique 64-bit GVA. 
As illustrated in Figure \ref{fig:gva}, the GVA space reserves two disjoint contiguous regions corresponding to HBM and DRAM, respectively.
Each region provides a GVA interval $[start, start + world\_size * max\_size]$, where $start$ is a globally agreed address base, $world\_size$ is the number of total processes, and $max\_size$ is the contributed maximum memory capacity. The address space is linearly divided among all processes and visible to all processes. The GVA segment owned by $i$-th process is $[start + i * max\_size, start + (i + 1) * max\_size)$. 
This abstraction provides a single, consistent memory view to all training processes, allowing transparent access to remote memory as if it were local.

A logical GVA is insufficient for direct data movement because training processes ultimately access memory through a physical address.
Similar to virtual memory management in modern operating systems, our pool maintains the page tables that map GVA used by processes to physical addresses in HBM and DRAM. 
The key difference is that the virtual address exposed to each process is no longer a small local VA region, but a globally consistent GVA region spanning pooled memory across nodes and memory tiers.
During initialization of the memory pool before training, each process reserves a contiguous GVA region from the segment $[start + i * max\_size, start + (i + 1) * max\_size)$, allocates physical memory from the local operating system, and binds it to the reserved GVA region, which may be less than $max\_size$.
Each process exports its memory mapping information to all other processes, which update their page table accordingly.
Then, each process has local access to the GVA and the page tables used for virtual address translation, eliminating distributed address resolution protocols on the hot path.


During training, the vision embeddings and text tokens are written into the GVA segment belonging to their own training process.
LLM consumes the constructed microbatches directly from GVAs without being aware of their actual storage location.
In the implementation layer, FlowTrain resolves the GVA to the corresponding physical address through the collected page tables.
The available transport engine is selected based on the source and destination location and memory types.
If the request crosses memory tiers within a node, FlowTrain invokes intra-node transfer engine like NVLink and HCCS. For cross-node destination, FlowTrain establishes (or reuses) transport endpoints and dispatches the transfer through an RDMA-class or other high-speed interconnect backend. The upper layers that operate on the GVA region are decoupled from hardware-specific addressing and transport details, enabling transparent memory access.

\subsection{Heterogeneous Parallel Allocator}\label{sec:allocator}
Existing resource allocation schemes \cite{zhangDistTrainAddressingModel2025} aim to optimize the worst pipeline stages, which cannot directly apply to decoupled dataflow.
After integrating the unified memory pool, the optimization objective for resource partition and parallelism strategy is essentially changed. 
The allocator aims to optimize the achievable throughput while avoiding producer-consumer imbalance under a fixed resource budget $R$.
Let $\mathcal{E}$ denote the set of encoder modules and $b$ denote the backbone. For each module $i \in \mathcal{E} \cup \{b\}$, $\mathcal{C}_i$ defines its set of feasible parallelism configurations. Each configuration $c \in \mathcal{C}_i$ has a throughput $\tau_i(c)$ and resource cost $r_i(c)$. We use the binary variable $x_{i,c}\in\{0,1\}$ to indicate whether $c$ is selected. The aggregated encoder production rate and backbone consumption rate can be defined as:

\vspace{-2.9mm}
\begin{small}
    \begin{equation}
  T_{\mathrm{enc}} = \sum_{e \in \mathcal{E}} \sum_{c \in \mathcal{C}_e} x_{e,c}\,\tau_e(c), \quad
  T_{\mathrm{bb}} = \sum_{c \in \mathcal{C}_b} x_{b,c}\,\tau_b(c)
\end{equation}
\end{small}

Then, flow matching problem is formalized as:

\vspace{-3mm}
\begin{small}
\begin{equation}
\label{equ:max}
\begin{aligned}
\operatorname{lexmax} \quad & (G,-F) \\
\text{s.t.}\quad
&G=\min(T_{\mathrm{enc}},\, T_{\mathrm{bb}}), \\
&F=\left|T_{\mathrm{enc}} - T_{\mathrm{bb}}\right|, \\
& \sum\nolimits_{c \in \mathcal{C}_i} x_{i,c} = 1,
\quad \forall i \in \mathcal{E} \cup \{b\}, \\
& \sum\nolimits_{i \in \mathcal{E} \cup \{b\}}
  \sum\nolimits_{c \in \mathcal{C}_i}
  x_{i,c}\,r_i(c) \le R.
\end{aligned}
\end{equation}
\end{small}
The objectives are optimized in lexicographic order. We first maximize the global throughput $G$, and then minimize flow imbalance $F$ between producers and consumers.
Over-allocating resources to the encoder incurs redundant buffering in the pool, while under-allocating starves the backbone.

However, it is intractable to directly solve the optimization problem in Eq. (\ref{equ:max}) for two reasons. 
On the one hand, the search space grows significantly with the cluster scale. For a cluster with 256 devices, the number of possible strategy combinations may exceed \(10^4\).
On the other hand, the steady-state throughput $\tau_i(c)$ depends on the combined effect of operator efficiency and communication overhead, neither of which can be explicitly formulated as a closed-form function.
To this end, the allocator exploits a three-step solution that prunes the search space, builds a throughput surrogate model, and solves the Mixed-Integer Linear Programming (MILP) problem.

\textit{Firstly}, we reduce the candidate space $\mathcal{C}_i, \forall i \in \mathcal{E} \cup \{b\}$ while retaining practically deployable configurations.
As the encoder is relatively small, we only consider DP and TP for the encoder side. Therefore, each candidate configuration occupies $\mathrm{DP}\times\mathrm{TP}$ devices. For the substantially larger backbone side, we additionally include PP, requiring $\mathrm{DP}\times\mathrm{TP}\times\mathrm{PP}$ devices per configuration. 
We prune invalid and low-efficiency configurations where the global batch size cannot be divisible by the resulting DP degree, and TP/PP degrees cannot factorize the per-module resource budget and cause cross-node tensor parallelism.
\textit{Secondly}, we perform throughput profiling by executing a small number of warm-up iterations under a fixed packed sequence length, which is the established methodology in parallelism planning \cite{um2024metis,miao2022galvatron}.
This lightweight profiling estimates a surrogate of $\tau_i(c)$ and is only performed once before VLM training begins, resulting in acceptable overhead compared to end-to-end training.
\textit{Thirdly}, substituting the surrogates into Eq. (\ref{equ:max}) yields the MILP problem, which can be efficiently solved by off-the-shelf solvers (e.g., Gurobi) in milliseconds.
Then, the obtained resource partition and module-specific parallelism strategies are materialized by Ray \cite{moritz2018ray} as a set of actors.

\subsection{Dynamic Packing Scheduler}\label{sec:scheduler}

\begin{figure}[tp]
\centering
\includegraphics[width=1.0\linewidth]{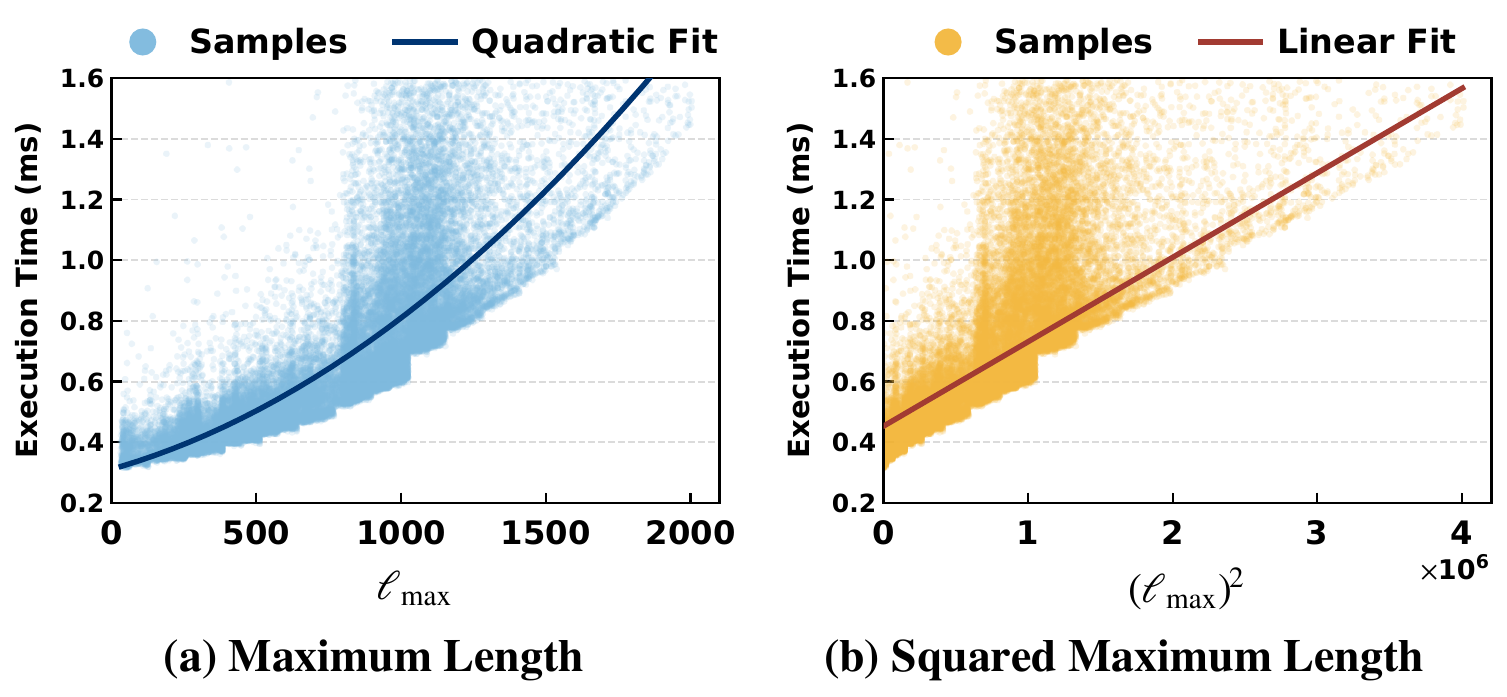}
\caption{Execution cost modeling with maximum sub-sequence length $\ell_{\max}$.}
\label{fig:max_len}
\vspace{-2mm}
\end{figure}

Similar to the allocator, the decoupled execution paradigm also reshapes the design space and scheduling objective. 
Regardless of monolithic or disaggregated design, microbatch composition is usually performed before multimodal encoding and only relies on raw sample size.
In comparison, our unified memory pool absorbs runtime variance, preventing it from propagating into the backbone.
The dynamic packing scheduler can flexibly assemble balanced microbatches from available embeddings in the pool based on the actual computational cost, rather than committing to fixed microbatch compositions before training begins. As summarized in Algorithm~\ref{alg:packing}, the scheduler minimizes padding waste through best-fit packing, and then relieves PP and DP imbalance via zig-zag dispatch.

\begin{algorithm}[h]
\caption{Dynamic Packing Algorithm}
\label{alg:packing}
\tcc{Best-fit Packing}
Initialize segment tree $\mathcal{T}$ over $[0, L]$\;
\For{Each sequence $s$ with length $\ell(s)$}{
  $j \leftarrow \textsc{BestFitQuery}(\mathcal{T}, \ell(s))$\;
  \eIf{$j \neq \texttt{null}$}{
    Append $s$ to microbatch $B_j$\;
    $P_j \leftarrow P_j - \ell(s)$ and update $\mathcal{T}$\;
  }{
    Create $B_{j}$ with $P_j \leftarrow L - \ell(s)$\;
    Insert into $\mathcal{T}$\;
  }
  \If{$P_j < \ell_{\min}$}{
    Remove $B_j$ from $\mathcal{T}$\; 
  }
}
\tcc{Two-dimension Load Balancing}
\For{Each packed microbatch $B_j$}{
  Compute $\mathrm{Cost}(B_j)$\;
  Assign $B_j$ to bucket
  $k$ by clustering $\mathrm{Cost}(B_j)$\;
}
\For{Each bucket $k$}{
  Obtain $list$ by sorting microbatches in descending order of $\mathrm{Cost}(B_j)$\;
Initialize $a \leftarrow 0$, \texttt{Forward}$ \leftarrow \text{True}$\;
    \While{$a < \text{len}(list)$}{
        \If{\texttt{Forward}}{
            \For{$n = 0$ \textbf{to} $N-1$}{
                Assign $list[a]$ to $\text{Rank}_{n}$\; 
                $a \leftarrow a + 1$\;
            }
        }
        \Else{
            \For{$n = N-1$ \textbf{to} $0$}{
                Assign $list[a]$ to $\text{Rank}_{n}$\;
                $a \leftarrow a + 1$\;
            }
        }
        $\texttt{Forward} \leftarrow \text{not } \texttt{Forward}$\;
    }
}
\end{algorithm}

We utilize the segment tree-based algorithm to pack variable-length sequences in fixed-capacity microbatches, thereby reducing memory fragmentation \cite{ding2024fewer}.
A segment tree covers the integer interval $[0, L]$, where $L$ is the maximum microbatch sequence length (e.g., 8192). Each leaf corresponds to the residual capacity of the candidate microbatches, and each internal node stores the maximum residual capacity in its subtree (Line 1).
For each incoming sequence $s$ with length $\ell(s)$, the scheduler first examines the left subtree, which represents smaller residual capacities. If its residual capacity is $\ge\ell(s)$, the search recursively descends into the left subtree. Otherwise, it proceeds to the right subtree. This process locates the open microbatch whose residual capacity is $\geq \ell(s)$ and minimal among all candidates by $O(\log L)$ best-fit queries (Lines 2-3).
If a suitable microbatch $B_j$ is found, append incoming sequence $s$ to $B_j$ and update its residual capacity to $P_j \leftarrow P_j - \ell(s)$ (Lines 4-6).
Otherwise, create a new microbatch with residual capacity $L - \ell(s)$ and insert it into the tree (Lines 7-9). When a microbatch's residual capacity falls below the minimum sequence length in the current distribution, it is removed from the segment tree (Lines 10-11).


Then, packed microbatches are assigned to PP stages and DP ranks to relieve imbalance across both dimensions.
To characterize execution costs, we profile packed microbatches with different sequence compositions.
We observe from Figure \ref{fig:max_len} that the execution time is approximately linear in $(\ell_{\max})^2$, where $\ell_{\max}=\max_{s \in B_j} \ell(s)$ is the maximum sub-sequence length within a packed microbatch $B_j$. This observation is consistent with the quadratic scaling of the attention kernel \cite{dao2024flashattention}. 
Meanwhile, the execution time of MLP layers scales approximately linearly with the total number of valid tokens.
Based on these observations, we estimate the execution cost of $B_j$ after maximum microbatch length ($L$) normalization:

\begin{small}
\begin{equation}
\label{eq:cost}
  \mathrm{Cost}(B_j) = \frac{1}{L}\sum_{s \in B_j} \ell(s)
    + \frac{1}{L^2}(\ell_{\max})^2
\end{equation}
\end{small}


Based on $\mathrm{Cost}(B_j)$ capturing both linear and quadratic computations, we cluster packed microbatches with similar computational intensities into a discrete bucket, ensuring consistent execution latencies across PP stages (Lines 12-14).
Within each bucket, the microbatches are sorted in descending order of $\mathrm{Cost}(B_j)$ (Lines 15-16).
The sorted microbatches are dispatched in order $\text{Rank}_0 \to \text{Rank}_{N-1}$ (Lines 17-22), and then the direction of the traversal is reversed to $\text{Rank}_{N-1} \to \text{Rank}_0$ (Lines 23-27). 
This zig-zag dispatch ensures that high and low-cost microbatches are paired on DP ranks, forcing the total cost to converge across the cluster \cite{abdelhamid2020prompt}. 

\section{Experiments}\label{sec:Evaluation}

\subsection{Experimental Setup}
\textbf{Models and Datasets.}
We adopt a representative VLM architecture, where the vision encoder adopts the ViT architecture \cite{dosovitskiy2020image}, and the LLM backbone is based on Qwen3 \cite{yang2025qwen3}. We train three model scales spanning small, medium, and large configurations, i.e., VLM-6B, VLM-32B, and VLM-72B. 
We evaluate FlowTrain on \textit{InfoVQA}~\cite{mathew2022infographicvqa}, an open-source visual question answering dataset, and \textit{Industrial} dataset collected from an e-commerce platform. 


\textbf{Training Configurations.}
All experiments are conducted on a production cluster with 256 NPUs (64GB) \cite{zhou2025accelerating}.
FlowTrain uses Ray to orchestrate heterogeneous actors and enable independent execution between the encoder and the LLM backbone. 
The image and text subsequences are interleaved to form sequences of up to 8192, and the global batch size is set to 256.





\begin{figure}[tp]
\centering
\includegraphics[width=1.0\linewidth]{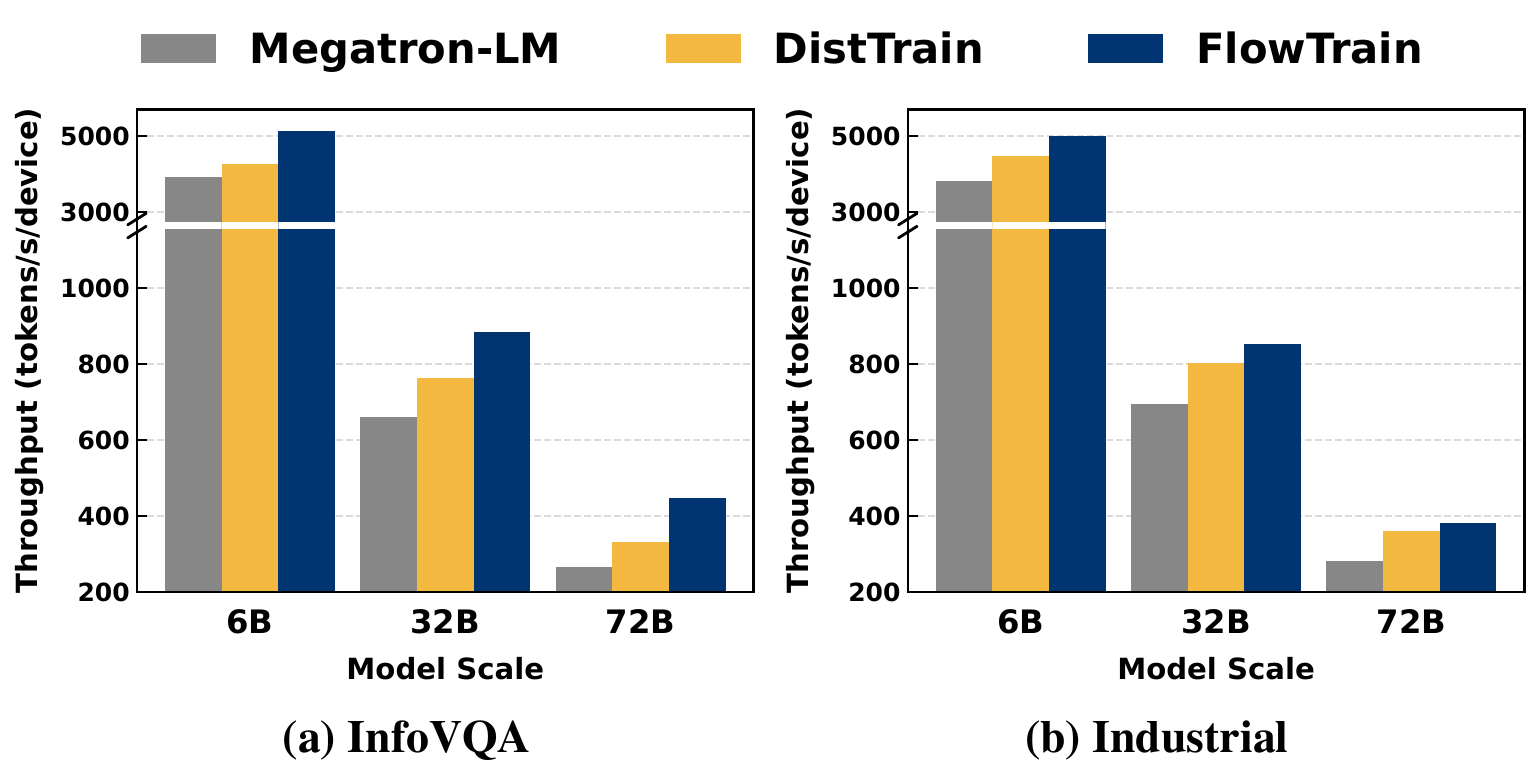}
\vspace{-6mm}
\caption{Training throughput across different methods on varying model sizes.}
\label{fig:diff_scale_model}
\vspace{-3mm}
\end{figure}

\textbf{Baselines.}
We compare FlowTrain with two representative baselines.
(1) \textit{Megatron-LM} \cite{shoeybi2019megatron} provides mature optimizations for LLM training and treats the multimodal model as a monolithic network.
(2) \textit{DistTrain} \cite{zhangDistTrainAddressingModel2025} adopts disaggregated model orchestration and data preprocessing, which deploy heterogeneous modules with independent resources and parallelism configurations.
We also report the LLM-only performance as a reference upper bound for unimodal training efficiency. 
For fair comparison, all methods are evaluated under the same software environment, optimizer configuration, and data pre-processing pipeline.


\textbf{Metrics.}
(1)~\textit{MFU} measures the fraction of hardware FLOPs converted into effective model computation, which directly quantifies resource efficiency independent of hardware count. (2)~\textit{Throughput} is reported as the number of tokens processed per second per device (tokens/s/device), reflecting training speed. (3)~\textit{Training loss} tracks the model's convergence trajectory throughout the training process, which is used to evaluate consistency.
(4)~\textit{Algorithm overhead} quantifies the average additional latency introduced by dynamic packing scheduling.

\begin{figure}[tp]
\centering
\includegraphics[width=1.0\linewidth]{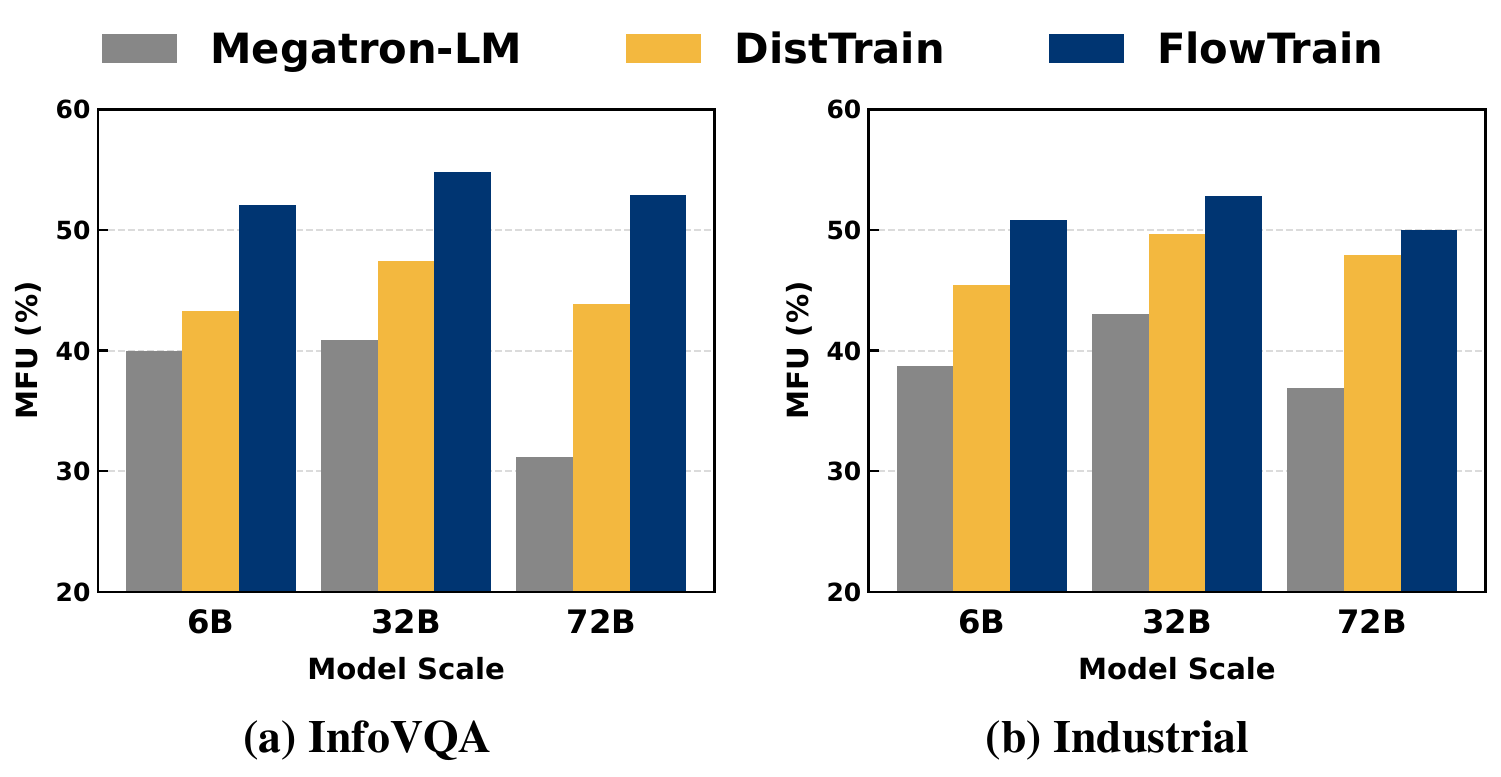}
\vspace{-6mm}
\caption{MFU across different methods on varying model sizes.}
\label{fig:two_dataset_mfu}
\vspace{-3mm}
\end{figure}

\subsection{End-to-end Effectiveness}
We evaluate the end-to-end training performance of FlowTrain with three model scales. 
The results in Figure~\ref{fig:diff_scale_model} show that FlowTrain consistently achieves the highest throughput among all training systems. 
FlowTrain achieves up to 1.7$\times$ and 1.4$\times$ throughput improvements on the InfoVQA dataset and the Industrial dataset, respectively.
Notably, the performance gap between FlowTrain and baselines increases as the models grow larger.
FlowTrain outperforms Megatron-LM by 30.4\% for VLM-6B, 33.7\% for VLM-32B, and 69.1\% for VLM-72B on the InfoVQA dataset.
This is because the baselines suffer from batch-level synchronization barriers, leading to poor hardware utilization.
In comparison, FlowTrain allows execution decoupling and coordinates heterogeneous modules through a shared memory abstraction, which substantially improves the training efficiency of large-scale VLMs.

\subsection{Resource Efficiency}
We compare the MFU of different methods with three model scales. 
As plotted in Figure~\ref{fig:two_dataset_mfu}, FlowTrain outperforms Megatron-LM by a large margin across all configurations.
Thanks to flow matching optimization between producers and consumers, FlowTrain achieves the MFU of 52.9\% and 50.0\% when training VLM-72B on the InfoVQA dataset and the Industrial dataset, while Megatron-LM only reaches 31.2\% and 36.9\% due to the monolithic design.
Besides, DistTrain partially alleviates the parallelism mismatch through disaggregated orchestration, but its MFU is still lower than that of FlowTrain. 
This gap is rooted in the rigid pipeline execution of DistTrain.
FlowTrain allows the LLM backbone to proceed without waiting for the entire batch of encoder output and to consume balanced microbatches, thus improving resource utilization.

\subsection{Scaling Performance}
\begin{figure}[tp]
\centering
\includegraphics[width=1.0\linewidth]{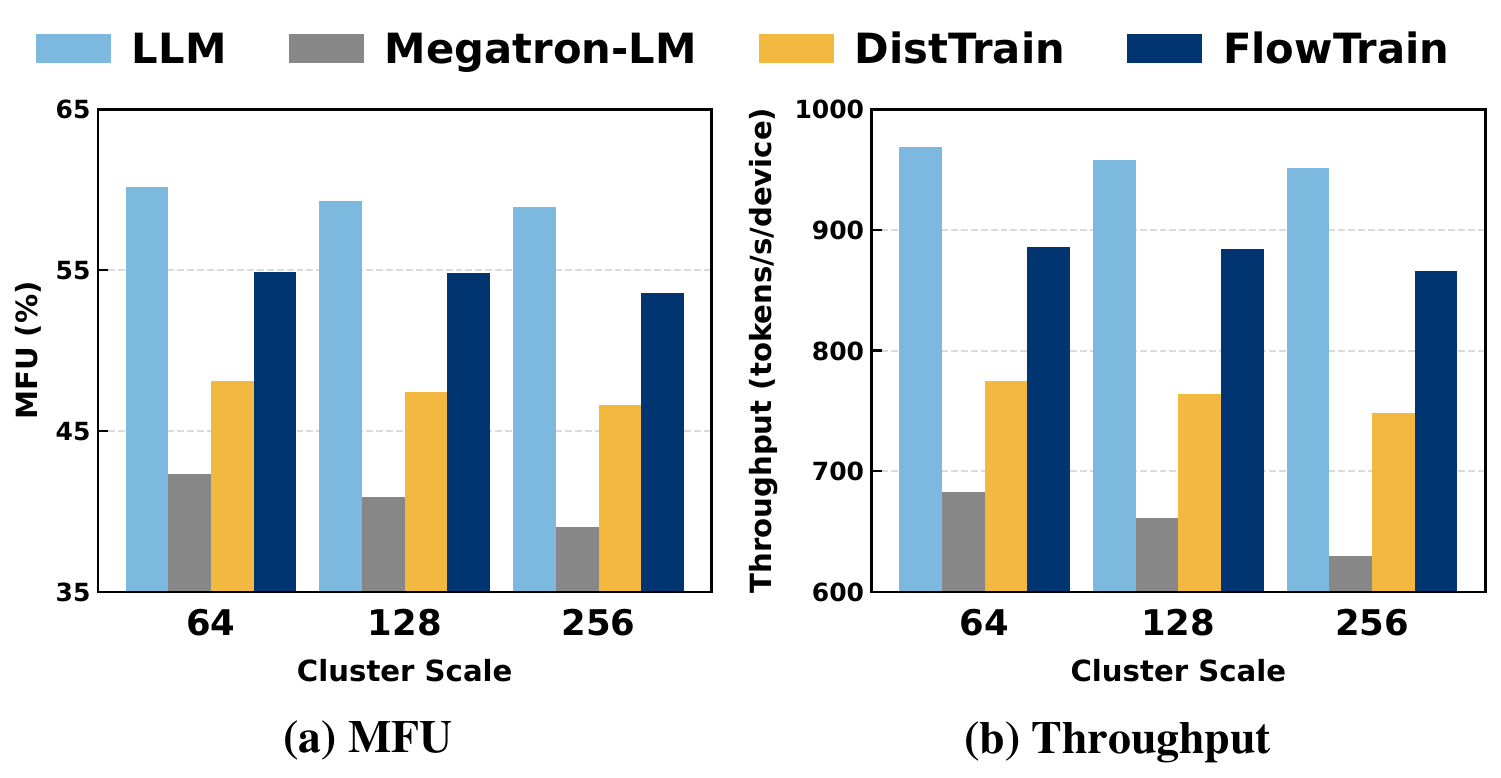}
\vspace{-6mm}
\caption{The efficiency gap compared to LLM-only training at different cluster scales.}
\label{fig:scaling}
\end{figure}

We further evaluate scalability by varying the cluster scales. 
Figure~\ref{fig:scaling} reports the results of training VLM-32B on the InfoVQA dataset, together with LLM-only training as a reference upper bound. 
FlowTrain achieves above 53\% MFU and 1.1-1.4$\times$ throughput improvement across all evaluation scales and exhibits the most robust scaling behavior, while the training efficiency of other methods is far lower than ours.
Another noteworthy observation is that FlowTrain is closer to LLM-only training performance than baselines.
The average MFU gaps are 18.7, 12.1, and 5.0 percentage points for Megatron-LM, DistTrain, and FlowTrain, respectively. 
These results confirm that our design can significantly narrow the gap to the efficiency regime associated with unimodal training workloads.

\begin{table}[tp]
\centering
\resizebox{0.5\textwidth}{!}{
\begin{tabular}{ c cccccc }
\toprule
\multirow{2}{*}{\textbf{Component}} & \multicolumn{2}{c}{\textbf{VLM-6B}} & \multicolumn{2}{c}{\textbf{VLM-32B}} & \multicolumn{2}{c}{\textbf{VLM-72B}} \\ 
\cmidrule(lr){2-3} \cmidrule(lr){4-5} \cmidrule(lr){6-7} 
& \textbf{MFU} & \textbf{Throughput}  & \textbf{MFU} & \textbf{Throughput}   & \textbf{MFU} & \textbf{Throughput} \\ \midrule

\textbf{w/o Pool}  & 21.6 &2118  &44.4    &717 &47.0 &399     \\
\textbf{w/o Allocator}  & 39.5 &3879  &50.1    &809 &48.9 &   415     \\
\textbf{w/o Scheduler}  & 46.1 &4531  &49.1    &793 &47.2 &   400      \\
\rowcolor{gray!20} \textbf{FlowTrain}  & 52.1 &5125  & 54.8    &884 &52.9 &448     \\

\bottomrule
\end{tabular}}
\caption{Ablation study of three core components on MFU (\%) and throughput (tokens/s/device).
}
\label{tab:Ablation}
\vspace{-2mm}
\end{table}

\subsection{Ablation Study}
We conduct ablation experiments to quantify the individual contribution of three components in FlowTrain. 
As summarized in Table \ref{tab:Ablation}, removing the unified memory pool causes the largest degradation, reducing MFU and throughput by 15.6 percentage points and 29.5\% on average, respectively.
The system falls back to point-to-point batch synchronization, which reintroduces idle time and weakens the main benefit of decoupled execution. 
Without the heterogeneous parallel allocator, the training throughput decreases by up to 24.3\%.
Due to throughput mismatch between producers and consumers, the backbone is either starved of embeddings or overwhelmed by excessive production in the pool.
Disabling the scheduler produces consistent performance loss.
The MFU drops to 46.1\%, 49.1\%, and 47.2\%, while the throughput decreases by 10.3-11.6\%.
The microbatches are no longer packed and dispatched according to actual computation cost, leading to padding waste and load imbalance. 
The ablation results demonstrate that three core components are essential and their combination delivers the maximum efficiency gain.

\begin{figure}[tp]
\centering
\includegraphics[width=1.0\linewidth]{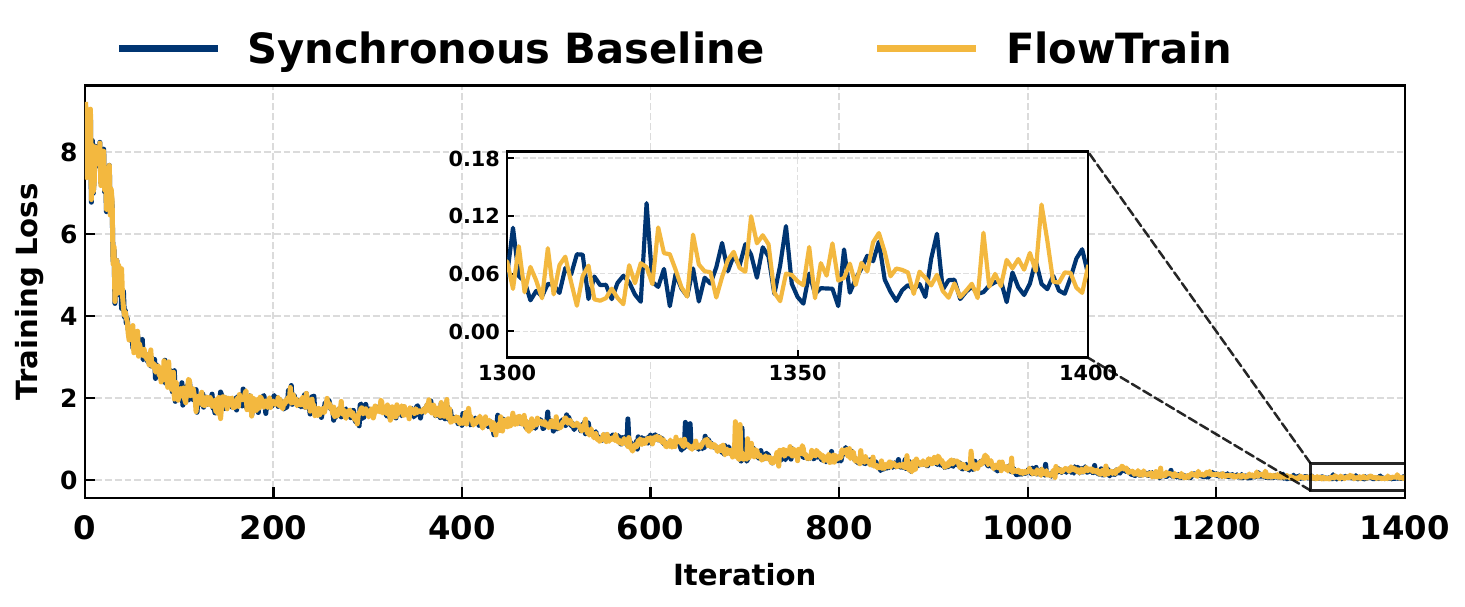}
\vspace{-7mm}
\caption{Training loss curves of FlowTrain and synchronous baseline.}
\label{fig:loss}
\vspace{-2mm}
\end{figure}

\subsection{Training Consistency}
We evaluate whether FlowTrain preserves the optimization behavior of synchronous training. 
Figure \ref{fig:loss} shows that FlowTrain closely follows synchronous training throughout the entire training process on the InfoVQA dataset.
Over the final 100 iterations, FlowTrain and baseline achieve average training losses of 0.0615 and 0.0553, respectively, corresponding to an absolute difference of only 0.0062.
The reason is that FlowTrain only changes the embedding execution and scheduling of frozen encoders but does not introduce stale activations or cross-version gradients. 
These results indicate that FlowTrain preserves synchronous updates and does not materially alter the observed optimization trajectory.

\subsection{Algorithm Overhead}
\begin{figure}[tp]
\centering
\includegraphics[width=1.0\linewidth]{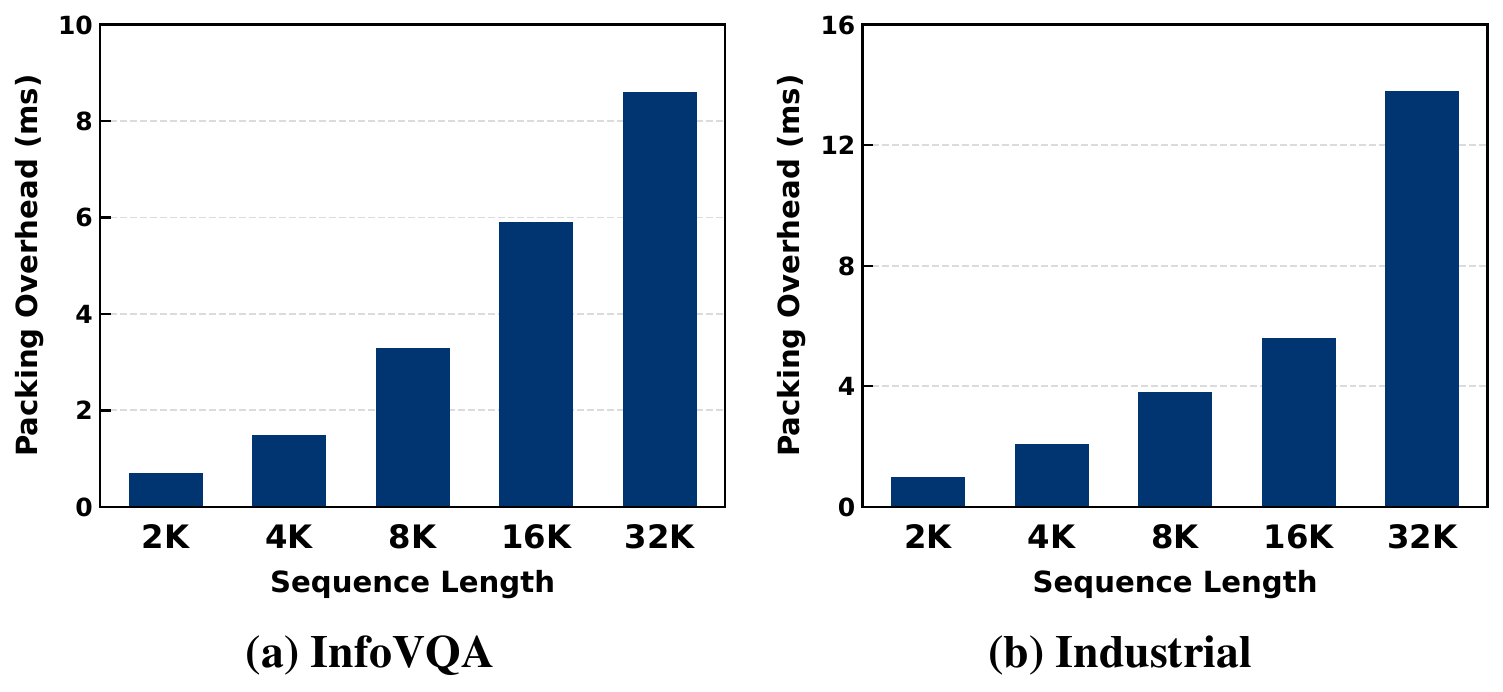}
\vspace{-6mm}
\caption{The average overhead of dynamic packing algorithm under different sequence lengths.}
\label{fig:overhead}
\vspace{-1mm}
\end{figure}

The additional overhead introduced by dynamic packing algorithm is a major concern in practical industrial deployments.
Figure \ref{fig:overhead} records that the average packing overhead for InfoVQA and Industrial datasets increases from 0.7-1.0ms to 8.6-13.8ms with sequence length varying from 2K to 32K. This upward trend is expected because longer sequences enlarge the candidate packing space and increase the cost of residual-capacity queries and load balancing. In particular, the Industrial dataset incurs larger algorithm overhead than the InfoVQA dataset due to its more severe data imbalance.
Nevertheless, the overhead remains small compared with the forward and backward computation time. Even at 32K length, the packing overhead accounts for less than 0.1\% of one training iteration.

\section{Related Work}
Compared with unimodal LLM training, optimizing VLM training is more difficult due to its inherent heterogeneity and dynamics. 
Driven by continuous scaling of industrial VLMs combined with trillion-level datasets, distributed training on large-scale clusters has become imperative \cite{dong2025scalable}.
Early systems such as DistMM \cite{huang2024distmm} can handle traditional multimodal training efficiently, but do not support LLM training naively.
Although frameworks like Megatron-LM \cite{shoeybi2019megatron,rasley2020deepspeed,zhao2023pytorch} excel at LLM workloads, they treat VLMs as monolithic models with heterogeneous layers, resulting in suboptimal resource allocation.

Other studies \cite{feng2025optimus,wang2025spindle,xue2025pipeweaver} adopt fine-grained bubble exploitation or pipeline scheduling algorithms to improve resource utilization.
Following this idea, LongCat-Flash-Omni~\cite{team2025longcat} and MegaScale-Omni \cite{xue2026megascale} further reduce engineering complexity. However, colocated schemes may suffer from the long-tail effect, resulting in resource conflicts and additional onload/offload overhead. 
Disaggregated systems represented by DistTrain \cite{zhangDistTrainAddressingModel2025} deploy heterogeneous modules on separate resource groups, addressing model and data heterogeneity.
However, these methods retain batch-level synchronization between heterogeneous modules, hindering efficient VLM training.
To fill this gap, FlowTrain enables disaggregated deployment and decoupled execution to improve resource utilization and training throughput.


\section{Conclusion}\label{sec:conclusion}

In this paper, we propose FlowTrain, a disaggregated and decoupled VLM training framework that breaks batch-level synchronization barriers through three core designs: a unified memory pool that enables flexible producer-consumer collaboration, a heterogeneous parallel allocator for throughput matching, and a dynamic packing scheduler constructing balanced microbatches.
Empirical evaluations on different models and scales demonstrate the efficiency of FlowTrain.

\bibliography{custom}

\appendix

\label{sec:appendix}

\end{document}